# A Meta-heuristically Approach of the Spatial Assignment Problem of Human Resources in Multi-sites Enterprise

* TKATEK Said
LaRIT, Faculty of Sciences
Ibn Tofail University
Kenitra, Morocco
saidtkinfo@yahoo.fr

ABDOUN Otman
Polydisciplinary Faculty,
Abdelmalk Essaadi University,
Larache, Morocco
otman.fpl@gmail.com

ABOUCHABAKA Jaafar
LaRIT, Faculty of Sciences,
IbnTofail University
Kenitra, Morocco
abouch06-univ@yahoo.fr

RAFALIA Najat
LaRIT, Faculty of Sciences,
IbnTofail University
Kenitra, Morocco
rafaliae@yahoo.com

## ABSTRACT
The aim of this work is to present a meta-heuristically approach of the spatial assignment problem of human resources in multi-sites enterprise. Usually, this problem consists to move employees from one site to another based on one or more criteria. Our goal in this new approach is to improve the quality of service and performance of all sites with maximizing an objective function under some managers imposed constraints. The formulation presented here of this problem coincides perfectly with a Combinatorial Optimization Problem (COP) which is in the most cases NP-hard to solve optimally. To avoid this difficulty, we have opted to use a meta-heuristic popular method, which is the genetic algorithm, to solve this problem in concrete cases. The results obtained have shown the effectiveness of our approach, which remains until now very costly in time. But the reduction of the time can be obtained by different ways that we plan to do in the next work.

## Keywords
Human Resources, Spatial Assignment Problem, Meta-heuristic Method, Genetic Algorithms, NP-hard Problem.

## 1. INTRODUCTION
The human resource management is one of the most critical problems and more costly for the majority of companies [5], and their problems occupy a place in company policies (public or private). The efficient and effective management of staff assignments is a key axis in the management of resources. It also imposes an essential tool to enhance profitability, increase productivity and improve business performance. Today, the real indicator to evaluate performance of these companies is no longer its financial availability but rather its ability to mobilize effective human resources, to assume the responsibilities and tasks assigned to them.

For this fact, companies seeking more and more a performance affirmed in a competitive environment, and they need more qualified human resources. However, the reassignment of certain skilled employees by the mobility process (spatial assignment process) to other places in the same company may be a key factor in the transfer of skills to the team that provides this site. This leads to improve the overall performance of the company [14].

Generally, the assignments of human resources can be set in two modes: Assignment for execute an activity without changing the workplace and assignment for execute an activity with changing workplace. This assignment mode requires an important tool for the optimization of human resources and remains a primary means to provide the adequacy among the requirements and resources, especially in large multi-site enterprises [9].

In this work, we define geographical mobility or movement of an employee to another location of the workspace by spatial mobility or spatial assignment that employee performs the same activity in the new workplace. This spatial mobility is designed in a rational approach and employees gradually binding decisions and accepts the changes necessary for the functioning of the company.

Many works have been developed in the field of human resource assignments in companies without addressing the problems of changing workplace in the same company. These studies have focused on the problems that primarily affect the human entity (assignment, planning and scheduling) in relation to the tasks to be executed and the constraints imposed by management. Among these works, we cite the work of Hachicha [15] proposed an approach to solve the problem of resource allocation with consideration of competence and preferences to constitute a system of correspondence between the achievements of an employee, the requirements of a task and the satisfaction of the performance of employers. The work of Bennour [5,6] was focused to propose method for optimizing the assignment of human resources in a enterprise, taking into account the performance and competencies in enterprise processes. The work of Campbell [7] proposes a model for maximizing the use of human resource skills for interdepartmental assignment and Hlaoitinun [14] proposed a method to support the formation of teams for coupling structuring project and piloting skills and other. Trojet work [22] has been focused on the planning of resources in a productive multi-site business, and proposed model as a problem of dynamic constraint satisfaction. In this work, we focus on studying a problem of spatial assignment in multi-sites enterprise in order to control and maximize the performance of sites. These types of companies are composed on Central office and Network of sites.

This paper is organized as follows: At first, we present a generalization of staff assignments by spatial mobility process with explaining the relationship between mobility and performance variation (overall performance caused by the spatial assignments or the spatial mobility). Subsequently, we present firstly the overall formulation of the problem. To avoid this difficulty, we have opted to use a meta-heuristic popular method, which is the genetic algorithm, to solve this problem in concrete cases. In the end, before the conclusion, we comment on and compare the results obtained by using summary tables and graphical results.

## 2. ASSIGNMENTS OF HUMAN RESOURCES BY SPATIAL MOBILITY

### 2.1 Definition of Spatial Mobility

Spatial mobility overlaps a rather broad dimension during the prospect of a change of assignment. This is one of the levers of human resources policy. The large multi-sites companies see the spatial mobility as a way to remove their human resources between the different sites by mobilizing their employees to different geographical locations to improve their competitiveness on the one hand and on the other hand their customer service [9].

### 2.2 Processes of Spatial Mobility

Usually spatial mobility involves managing the business movement between sites of a company by a given process: Assignment, reassignment or redeployment.

*The Assignment process:* is the set of consecutive procedures of recruiting, and then assign an employee belonging to a category and having a specialty. The employee must be qualified, competent and dedicated to assign it to a company site.

*The reassignment process:* is the set of consecutive procedures of moving an employee within the company to assuming the role of an operational position to complete a task.

*The redeployment process:* is the set of consecutive procedures of moving an employee to another site in order to provide quality human resources, able to cover the needs voiced in this site.

So this problem is to mobilize, between different sites, employees belonging to a class of staff and having a specialty with one of the processes described above. The two parameters require that the employee perform a well-defined activity. This type of assignment has a choice destination site, evaluation criteria and constraints imposed by managers.

*Choice of destination site*: is that an employee of a category and a specialty chooses one or more allocation sites to perform the same activity.

*Evaluation criteria*: have been the characteristics from which managers can evaluate the behavior of an employee as professional competence, performance, efficiency, effectiveness.

We call the value associated with human characteristics by performance. In this work, we combine the value of this performance in one parameter noted the weight *W*.

The overall weight is calculated, by using an employee evaluation function noted $W(X)=\sum \alpha_l w_l(X)$, with $\alpha_l$ is a weighting coefficient associated to the criteria $w_l(X)$ of a employee *X*. These criteria have professional nature (seniority in the post), social (nearby), productive (productivity and profitability).

## 3. POSITIONING, FORMULATION AND COMPLEXITY OF THE PROBLEM

### 3.1 Position of the Problem

In the problem of assignment of human resources through spatial mobility we concentrate primarily on the following criteria:

- Candidates are in the same professional class;
- Candidates in the same business;
- Candidates can choose only one destination site.

Taking into account these criteria, we are interested in formalizing the problem by determining the objective function. The latter will be maximized in terms of total weight while satisfying some constraints.

### 3.2 Formulation of the Assignment Problem by Spatial Mobility

In general, this problem can be stated as follows:
We have *NS* sites (sub-departments, subsidiaries, sites) $A_1,..., A_j, .., A_k,, A_{NS}$ contain candidates. The problem is how to choose candidates located in site $A_j$ and wanting to move to site $A^{(k)}$ with ($j \neq k$) to maximize the total weight (quality of service or overall performance) induced by spatial mobility of candidates, while respecting all the constraints imposed by manager. To answer to this question, we will present the formalism governing all assignments candidates by spatial mobility under these constraints.

For this, we first reorder the elements of the set $A_j^{(k)}$ using the weight function *W(X)* as follows: $\forall\ i \in [1, \tilde{N}_{jk}]$ with $\tilde{N}_{jk}$ is the number of candidates of site $A_j$ wishing to mobilize to site $A^{(k)}$, we have : $W_{(i+1)j}^k > W_{ij}^k$ *(j, k)* $\in [1, NS]^2$ , with $W_{ij}^k$ : is the weight measured of the candidate *i* located in site $A_j$ and wanting to move to site $A^{(k)}$.

Let $X_{ij}^k$ a decision variable such that:

$$X_{ij}^k = \begin{cases} 1 & \text{If the candidate } i \text{ of site } A_j \text{ is assigned to site } A^{(k)} \\ 0 & \text{Otherwise} \end{cases}$$

So, the number of candidates who will be mobilized to site $A^{(k)}$ is given by:

$$N_k = \sum_{\substack{j=1 \\ j \neq k}}^{NS} \sum_{i=1}^{\tilde{N}_{jk}} X_{ij}^k \qquad (1)$$

Let $W_{glo}^k$ the total weight induced during displacement candidates having an individual weight $W_{ij}^k$, this total weight is given by:

$$W_{glo}^k = \sum_{\substack{j=1 \\ j \neq k}}^{NS} \sum_{i=1}^{\widetilde{N}_{jk}} W_{ij}^k X_{ij}^k \quad (2)$$

We construct a matrix $\beta^k$ such that :

$\beta^k = \{\beta_{ji}^k / 1 \leq j \leq NS;\ 1 \leq i \leq max(\widetilde{N}_{jk}); j \neq k\}$ With:

- The Site $A_j$ is represented by a column vector *j*.
- The candidate *i* located in site $A_j$ is represented by a row.
- $\beta_{ji}^k = W_{ij}^k$ is the element of matrix $\beta^k$.

Similarly, we construct a binary matrix $X^k$ such that:

$X^k = \{X_{ij}^k / 1 \leq i \leq NS;\ 1 \leq j \leq max(\widetilde{N}_{jk}); j \neq k\}$

Global matrix $\beta$ weight and global assignment *X* are respectively formed by the sub matrix $\beta^k$ and $X^k$ :

$$\beta = (\beta^k), \ X = (X^k) \text{ with } 1 \leq k \leq NS$$

The optimization problem assignments spatial mobility of candidates likely to be mobilized to sites $A^{(k)}$ is to determine the different decisions $X = \{X^k\}$ leading to the maximization of the overall weight. The objective function associated to this problem can be written as:

$$F(X) = \sum_{k=1}^{NS} |\sum_{\substack{j=1 \\ j \neq k}}^{NS} \sum_{i=1}^{\widetilde{N}_{jk}} \beta_{ji}^k X_{ij}^k| = \sum_{k=1}^{NS} |trace(\beta^k X^k)| \quad (3)$$

If we set $F_k(X^k) = trace(\beta^k X^k)$

This objective function is optimized as follows:

$$Max\ (F(X)) = Max\ (\sum_{k=1}^{NS} |F_k(X^k)|) \quad (4)$$

## 3.3 Constraints of the Problem

### 3.3.1. Constraints of capacity

This constraint denote that the number of candidates who will be moved to the destination site $A^{(k)}$ must be equal to one expressed by the leaders.
So the overall weight matrix β and the assignment matrix X are formed respectively by sub matrix $\beta^k$ and $X^k$.

$$\beta = \begin{pmatrix} \beta^{k=1} \\ \beta^{k=2} \\ \beta^{k=3} \end{pmatrix} \text{ and } X = \begin{pmatrix} X^{k=1} \\ X^{k=2} \\ X^{k=3} \end{pmatrix}$$

### 3.3.2. Global formulation of the problem

From equations (4), (5), (6) and (7) we deduce the global formulation of the problem of spatial mobility assignments which can be written as:

$$\begin{cases} Max(F(x)) = Max \sum_{k=1}^{NS} (|Trace(\beta^k X^k)|) \\ SC\ \ ||X^k||_1 \leq C_k \quad \forall\ k\ \in [1, NS] \\ SC\ \ X_{lj}^k \left(\sum_{i=1}^{l} X_{ij}^k - l\right) = 0\ \forall\ l \in [1, \widetilde{N}_{jk}] \quad (8) \\ SC\ \sum_{k=1}^{NS} X_{ij}^k \leq 1\ \ \forall\ i \in [1, \widetilde{N}_{jk}]\ , \forall\ j \in [1, NS] \end{cases}$$

### 3.3.3. Complexity of the problem

The complexity of this problem is very important and classified as NP-hard problems because the formulation presented coincides perfectly with a combinatorial optimization problem (COP), and also seen as a dynamic problem. Therefore, we turned to a heuristic method using an iterative local search. The complexities of the search space and function to maximize use methods lead to radically different resolutions [21, 23]. In most cases, an optimization problem is naturally divided into two phases: research feasible solutions and to search for the optimal solution among these solutions. Depending on the method used, this division is more or less apparent in the resolution. The use of a genetic algorithm [10, 14] is suitable for rapid and comprehensive exploration of a large space search and is able to provide several solutions and choose the best among these solutions.

## 4. GENETIC ALGORITHM

The genetic algorithm is a one of the family of evolutionary algorithms. The population of a genetic algorithm evolves by using genetic operators inspired by the evolutionary in biology[1]. In Genetic Algorithm, a population of potential solutions termed as chromosomes and individuals are evolved over successive generations using a set of genetic operators called selection, crossover and mutation. First of all, based on some criteria, every chromosome is assigned a fitness value and then the selection operator is applied to choose relatively fit chromosomes to be part of the reproduction process. In reproduction process new individuals are created through application of operators. Large number of operators has been developed for improving the performance of GA, because the performance of algorithm depends on the ability of these operators [2]. The mutation and crossover operator is used to maintain adequate diversity in the population of chromosomes and avoid premature convergence.

These algorithms were modeled on the natural evolution of species. We add to these evolution concepts the observed properties of genetics (Selection, Crossover, Mutation, ..) (Fig. 1). They attracted the interest of many researchers, starting with Holland [19], who developed the basic principles of genetic algorithm, and Goldberg [11] has used these principles to solve a specific optimization problems. Other researchers have followed :

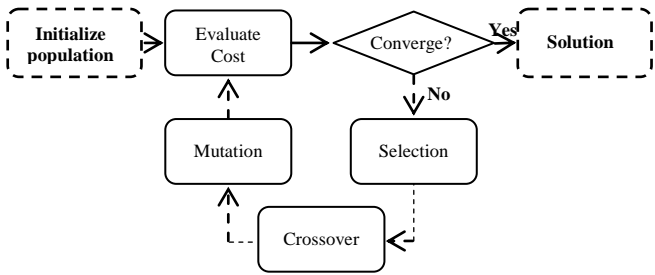

**Fig 1: Flowchart of optimization with a genetic algorithm [3]**

## 4.1 Advantages

Compared to the classical optimization algorithms, the genetic algorithm has several advantages as [1]:

- Use only the evaluation of the objective function regardless of its nature. In fact we do not require any special property of the function to be optimized (continuity, differentiability, connectedness, ..), which gives it more flexibility and a wide range of applications;
- Generation has a parallel form by working on several points at once (population of size N) instead of a single iteration in the classical algorithms;
- The use of probabilistic transition rules (crossover and mutation probability), as opposed to deterministic algorithms where the transition between two individuals is required by the structure and nature of the algorithm.

## 4.2 Principles and Functioning

The genetic algorithms are based on six principles [1, 3]:

- Each treated problem has a specific way to encode the individuals of the genetic population. A chromosome (a particular solution) has different ways of being coded: numeric, symbolic, or alphanumeric;
- Creation of an initial population formed by a finite number of solutions;
- Definition of an evaluation function (fitness) to evaluate a solution;
- Selection mechanism to generate new solutions, used to identify individuals in a population that could be crossed, there are several methods in the literature, citing the method of selection by rank, roulette, by tournament, random selection, etc.;
- Reproduce the new individuals by using Genetic operators:
  i. *Crossover operator*: is a genetic operator that combines two chromosomes (parents) to produce a new chromosome (children) with crossover probability $P_x$;
  ii. *Mutation operator*: it avoids establishing it avoids population unable to evolve. This operator used to modify the genes of a chromosome selected with a mutation probability $P_m$;

- Insertion mechanism: to decide who should stay and who should disappear.
- Stopping test: to make sure about the optimality of the solution obtained by the genetic algorithm.

In the following, we apply a genetic resolution in a typical example in order to obtain an optimal solution from several solutions generated verifying the objective function of the overall weight induced and the values of the constraints imposed.

## 5. RESULTS AND ANALYSIS
## 5.1 Description of a Sample Test

To carry out this study and evaluate this work, we deal with an example of medium size. We believe that a company is formed by three sites $A_1$, $A_2$ and $A_3$. The brings in a total number of candidates equal to 29 candidates, each of these candidates has a $W_i$ weight that characterizes his professional assessment (competence, productivity, profitability ...) wishing to affect or move by spatial mobility to another site.

Tables 1, 2 and 3 summarize the values of the weights of the candidates as well as sites of origin and destination $A_j$ and $A^{(k)}$.

**Table 1. Destination individuals to $A_1$ (k = 1)**

| Original site | Destination site | Weight (W) |
|---|---|---|
| A2 | A1 | 76 |
| A2 | A1 | 67 |
| A2 | A1 | 43 |
| A2 | A1 | 43 |
| A2 | A1 | 29 |
| A2 | A1 | 22 |
| A2 | A1 | 20 |
| A2 | A1 | 16 |
| A2 | A1 | 13 |
| A3 | A1 | 80 |
| A3 | A1 | 41 |

**Table 2. Destination individuals to $A_2$ (k = 2)**

| Original site | Destination site | Weight (W) |
|---|---|---|
| A1 | A2 | 83 |
| A1 | A2 | 78 |
| A1 | A2 | 76 |
| A1 | A2 | 69 |
| A1 | A2 | 58 |
| A1 | A2 | 21 |
| A1 | A2 | 9 |
| A3 | A2 | 86 |
| A3 | A2 | 74 |
| A3 | A2 | 60 |
| A3 | A2 | 57 |
| A3 | A2 | 39 |

**Table 3. Destination individuals to $A_3$ (k = 3)**

| Original site | Destination site | Weight (W) |
|---|---|---|
| A1 | A3 | 64 |
| A1 | A3 | 27 |
| A1 | A3 | 7 |
| A2 | A3 | 67 |
| A2 | A3 | 11 |
| A2 | A3 | 4 |

**Table 4. The constraint values of $C_k$**

| $C_k$ | $C_k(k=1)$ | $C_k(k=2)$ | $C_k(k=3)$ |
|---|---|---|---|
| Constraint | 7 | 7 | 5 |

## 5.2 Synthetic Table of Results

In this section, we present a concrete example of this meta-heuristic approach proposed. The code for solve the problem of spatial mobility in this test has been developed by the C++ language.

In this example, the initial population contains 30 individuals and the stop condition depending on the convergence to the optimal solution of this problem. The results are summarized in Table 5, they were obtained on a Pentium 4 at 2 GHz with 1GB of RAM. The global weight matrix β used in this test is given by:

$$\beta = \begin{pmatrix} \beta 1 \\ \beta 2 \\ \beta 3 \end{pmatrix}$$

With :

$$\beta 1 = \begin{pmatrix} 76 & 67 & 43 & 43 & 29 & 22 & 20 & 16 & 13 \\ 80 & 54 & 0 & 0 & 0 & 0 & 0 & 0 & 0 \end{pmatrix}$$

$$\beta 2 = \begin{pmatrix} 83 & 78 & 76 & 69 & 58 & 21 & 9 \\ 86 & 87 & 60 & 57 & 39 & 0 & 0 \end{pmatrix}$$

$$\beta 3 = \begin{pmatrix} 64 & 27 & 7 \\ 67 & 11 & 24 \end{pmatrix}$$

The solution *X* of the problem is obtained as a matrix. This matrix is called global assignment matrix that is written as:

$$X = \begin{pmatrix} X1 \\ X2 \\ X3 \end{pmatrix}$$

## 5.3 Generating all Solutions of the Problem

The global assignment matrix with the process of spatial mobility is a binary matrix; the value 1 indicates that the candidate is assigned. The value 0 indicates that the candidate is not affected or mobilized.

This decision variable $X_{ij}^k$ ($j \neq k$) is assigned to i-th candidate of site $A_j$ wanted to mobilize to site $A^{(k)}$. This candidate is identified by $A_{ij}^{(k)}$ ($j \neq k$).

**Table 5. Solutions depending on the values of $F(X) = \text{Trace}(\beta X)$**

| Solution Si | Trace( βX) | Solution Si | Trace( βX) |
|---|---|---|---|
| S1 | 751 | S16 | 944 |
| S2 | 763 | S17 | 950 |
| S3 | 778 | S18 | 950 |
| S4 | 794 | S19 | 961 |
| S5 | 821 | S20 | 974 |
| S6 | 832 | S21 | 985 |
| S7 | 837 | S22 | 993 |
| S8 | 864 | S23 | 1003 |
| S9 | 875 | S24 | 1004 |
| S10 | 890 | S25 | 1014 |
| S11 | 905 | S26 | 1034 |
| S12 | 906 | **S27** | **1036** |
| S13 | 907 | **S28** | **1045** |
| S14 | 933 | **S29** | **1074** |
| S15 | 934 | **S30** | **1081** |

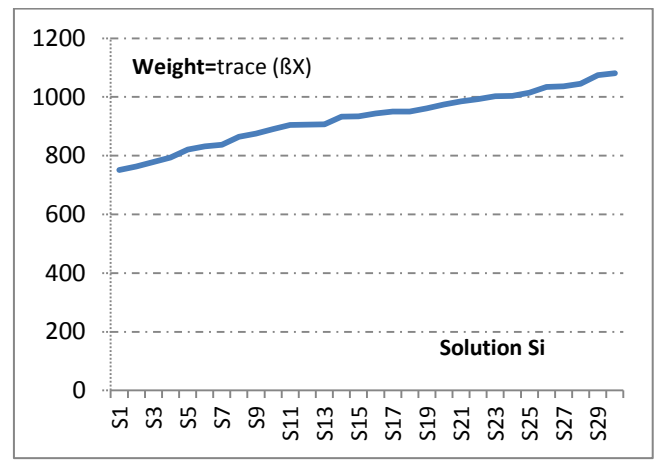

**Fig 2: The global weight induced depending on solutions obtained**

We have 30 different solutions obtained by performing 30 iterations, these solutions are ranked from 1 to 30 according to a decreasing order of the value associated with *F (X) = Trace (βX)*.

| **SOLUTION 30** <br> *trace (ßX) = 1081* <br><br> Sub Matrix X1 <br> 1 1 1 1 1 0 0 0 0 <br> 1 1 <br> Sub Matrix X2 <br> 1 1 1 1 0 0 0 <br> 1 1 1 0 0 <br> Sub Matrix X3 <br> 1 1 1 <br> 1 1 0 | **SOLUTION 29** <br> *trace (ßX) = 1074* <br><br> Sub Matrix X1 <br> 1 1 1 1 1 0 0 0 0 <br> 1 1 <br> Sub Matrix X2 <br> 1 1 1 1 0 0 0 <br> 1 1 1 0 0 <br> Sub Matrix X3 <br> 1 1 0 <br> 1 1 0 |
|---|---|
| **SOLUTION 28** <br> *trace (ßX) = 1045* <br><br> Sub Matrix X1 <br> 1 1 1 1 0 0 0 0 0 <br> 1 1 <br> Sub Matrix X2 <br> 1 1 1 1 0 0 0 <br> 1 1 1 0 0 <br> Sub Matrix X3 <br> 1 1 0 <br> 1 1 0 | **SOLUTION 27** <br> *trace (ßX) = 1014* <br><br> Sub Matrix X1 <br> 1 1 1 1 1 0 0 0 0 <br> 1 1 <br> Sub Matrix X2 <br> 1 1 1 1 0 0 0 <br> 1 1 0 0 0 <br> Sub Matrix X3 <br> 1 1 0 <br> 1 1 0 |

**Fig 3: Example of first solutions, depending on the value descendent of *trace(ßX)***

## 5.4 Selection of the Optimum Solution

In the 30 solutions generated by this approach, we choose the solution that maximizes the objective function while satisfying all the constraints. The optimal solution corresponds to the maximization of the objective function of the total weight F (X) = Trace (ßX) = 1081.

$$X = \begin{pmatrix} X1 \\ X2 \\ X3 \end{pmatrix}$$

With :

$$X1 = \begin{pmatrix} 1\,1\,1\,1\,1\,0\,0\,0\,0 \\ 1\,1\,0\,0\,0\,0\,0\,0\,0 \end{pmatrix}$$

$$X2 = \begin{pmatrix} 1\,1\,1\,1\,0\,0\,0 \\ 1\,1\,1\,0\,0\,0\,0 \end{pmatrix}$$

$$X3 = \begin{pmatrix} 1\,1\,1 \\ 1\,1\,0 \end{pmatrix}$$

The global matrix *X* assignment of spatial mobility contains three sub-matrix *X1*, *X2* and *X3*, the values of one sub matrix *Xi (i = 1, 2* or *3)* shows all possible assignments of candidates produced by spatial mobility for the best overall weight induced by these assignments. By analyzing these, we deduce the following:

The sub matrix *X1*, we have 5 candidates in the site $A_2$ are assigned to the site $A_1$ (row 1) and 2 candidates from the site $A_3$ are assigned to the site $A_1$ (row 2). In addition, the number of candidates who have left the site $A_1$ is 7. Same scenario for has been applied in the sub matrix X2 and X3.

The following table 6 summarizes all of the assignments made by the spatial mobility:

**Table 6. Summary of all possible assignments by spatial mobility**

| Site $A_j$ | Site $A^{(k)}$ | Number of candidates assigned to $A^{(k)}$ | Number of candidates out from $A^{(k)}$ |
|---|---|---|---|
| $A_2$ | $A^{(1)}$ | 5 | 7 |
| $A_3$ | $A^{(1)}$ | 2 | |
| $A_1$ | $A^{(2)}$ | 4 | 7 |
| $A_3$ | $A^{(2)}$ | 3 | |
| $A_1$ | $A^{(3)}$ | 3 | 5 |
| $A_2$ | $A^{(3)}$ | 2 | |

From the assignment matrix obtained, if we remove the various possible transpositions between candidates mobilized we find that three of candidates still where each is belongs to a site. These three nodes A1, A2 and A3 sites are formed a closed cycle.

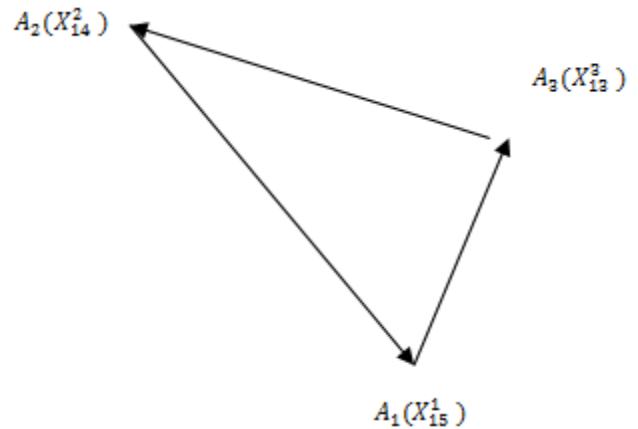

**Fig 4: Directed graph contains a closed cycle**

## 6. CONCLUSION

We have presented and evaluated in this paper a problem of assignment by the spatial mobility of human resources in a multi-sites enterprise. Firstly, we have presented global formulation of the optimization problem by including several quantitative and qualitative constraints (capacity, priority anduniqueness).Secondly; we used genetic algorithms to find solution of this problem. The results of test showed the importance of this work to control and improve service quality and performance of multi-site enterprises by reallocation of candidates by using spatial mobility.